\documentclass[]{interact}

\usepackage{epstopdf}% To incorporate .eps illustrations using PDFLaTeX, etc.
\usepackage[caption=false]{subfig}% Support for small, `sub' figures and tables
\usepackage{amsmath}
\usepackage{graphicx}
\usepackage{enumerate}
\usepackage{url} % not crucial - just used below for the URL 
\usepackage{bm}
\usepackage{amssymb} % has symbol for reals
\usepackage{mathtools} % has a big cartesian product symbol
\usepackage{setspace}
\usepackage{algorithm}
\usepackage{algorithmic}
\usepackage{xspace}

\usepackage{multirow}

\usepackage[table,xcdraw]{xcolor}
\usepackage{adjustbox}
\usepackage{lineno}
\usepackage{pdflscape}
\usepackage[numbers,sort&compress]{natbib}% Citation support using natbib.sty
\bibpunct[, ]{[}{]}{,}{n}{,}{,}% Citation support using natbib.sty
\renewcommand\bibfont{\fontsize{10}{12}\selectfont}% Bibliography support using natbib.sty
\makeatletter% @ becomes a letter
\def\NAT@def@citea{\def\@citea{\NAT@separator}}% Suppress spaces between citations using natbib.sty
\makeatother% @ becomes a symbol again

\theoremstyle{plain}% Theorem-like structures provided by amsthm.sty

\theoremstyle{definition}

\theoremstyle{remark}

\DeclareMathOperator*{\argmin}{argmin}

\newcommand{\eg}{e.g.\@ }
\newcommand{\ie}{i.e.\@ }

\begin{document}

%\articletype{ARTICLE TEMPLATE}% Specify the article type or omit as appropriate

\title{Generating Exact Optimal Designs via Particle Swarm Optimization: Assessing  Efficacy \textit{and} Efficiency via Case Study}

\author{
\name{Stephen J. Walsh\textsuperscript{a}\thanks{CONTACT Stephen J. Walsh. Email: steve.walsh@usu.edu} and John J. Borkowski\textsuperscript{b}}
\affil{\textsuperscript{a}Utah State University, Logan, Utah; \textsuperscript{b}Montana State University, Bozeman, Montana}
}

\maketitle

\begin{abstract}
In this study we address existing deficiencies in the literature on applications of Particle Swarm Optimization to generate optimal designs. We present the results of a large computer study in which we bench-mark both efficiency and efficacy of PSO to generate high quality candidate designs for small-exact response surface scenarios commonly encountered by industrial practitioners. A preferred version of PSO is demonstrated and recommended. Further, in contrast to popular local optimizers such as the coordinate exchange, PSO is demonstrated to, even in a single run, generate highly efficient designs with large probability at small computing cost. Therefore, it appears beneficial for more practitioners to adopt and use PSO as tool for generating candidate experimental designs. 
\end{abstract}


\section{Introduction}
A recent paper by Jensen (2018) highlights the growing popularity of optimal design of experiments (DoE) in industrial and manufacturing applications and identifies the need for continued research on this problem \cite{jensen1}. Generating optimal designs is well-recognized as the difficult optimization of a non-convex (specifically multimodal) criterion. The recent shift toward optimal design is due, in part, to the contemporary availability of computing and algorithms that are effective at generating candidate optimal designs. In this regard the coordinate exchange (CEXCH) of Meyer and Nachtsheim (1995) appears to be the most popular algorithm for generating optimal designs in both application and research \cite{cexch, odt}. However, CEXCH is well-recognized to be a local optimizer, and therefore several authors recommend applying CEXCH thousands of times and inspecting the quality of the resulting solutions except in the most simple design scenarios \cite{odt}. To address this issue the class of meta-heuristic evolutionary algorithms, \eg genetic algorithms (GA), simulated annealing (SA), and particle swarm optimization (PSO) have proven to be superior at finding designs with high optimality, albeit often at significant time and computing cost \cite{jobo1, licost, wong, shi1, ahdopt, gIopt}. Meta-heuristics are attractive to the optimal design problem because 1) they make little to no assumptions regarding the nature of the objective (optimality criterion) and 2) as opposed to CEXCH, meta-heuristics attempt to search the space of candidate design matrices globally and are robust to entrapment in local optima thereby giving them a significant advantage over CEXCH to generate a highly efficient optimal design in a single run.

Among the meta-heuristics, PSO is a relative newcomer to optimal design, with the first such application in 2013 by Chen et. al. which discussed PSO-generated latin-hypercube designs \cite{chen2}. Application of PSO to generating space-filling designs is addressed by \citep{chen1, joseph}. PSO for generating optimal designs for non-linear models is illustrated in \citep{lukemire, chen3, chen4}. PSO-generation of Bayesian continuous designs is discussed in \cite{shi1} and PSO for constructing continuous optimal designs for mixture experiments is provided in \cite{wong}. Therefore, the literature is lacking application of PSO to generating exact optimal designs and specifically to response surface settings and optimality criterion commonly employed in industrial applications. This is one contribution of this paper. 

The aforementioned papers which apply PSO to generate optimal designs only address algorithm efficacy \cite{chen2, chen1, joseph, lukemire, chen3, chen4, shi1, wong}. That is, they have a tendency to apply PSO to a problem and either show that it was able to produce a design with commensurate or better quality than those known, or to high-quality designs generated by other algorithms. Further, they only apply the simplest version of PSO which employes a global communication topology whereas the PSO literature suggests a local communication topology may be more effective for optimizing multimodal objectives such as the exact optimal design criteria \cite{spso1_2007, spso2_2011}. In addition, the exact design problem has no general theory (for example as continuous designs have the General Equivalence Theorem) with which to assess the quality of a proposed design. Therefore, it is necessary to generate exact designs under repeated application of the algorithm (which is recommended for the CEXCH \cite{odt}) in order to assess the efficiency of an algorithm for generating high-quality exact optimal designs. Such a replication study, which includes two PSO-variants which investigates different particle communication topologies, is a second contribution of this paper. Further, the need for such detailed benchmarking studies is consistent with the implications of the ``No Free Lunch'' theorems of Wolpert and Macready \cite{nofl} and in-line with commonly employed studies of benchmarking performance of meta-heuristics which are common in the computer science literature \eg \cite{spso2_2011}. 

In section 2 we present the notation of response surface designs and define the $D$- and $I$-optimal design criteria. In section 3 we describe the PSO algorithm and provide an extension of PSO to the optimal design problem. In section 4 we describe the large collection of computer experiments implemented to benchmark PSO performance and to study the effects of the following factors on the difficulty of the exact optimal design problem: $K$ (number of design factors), $N$ (number of design points), $S$ (number of particles in a swarm for a single PSO search, \ie this is the number of candidate optimal designs), PSO-version with levels Basic PSO (which utilizes the commonly used global communication topology) and SPSO2007 (which utilizes the local communication topology), and objective-class with levels $D$- and $I$-criterion. We provide the results in section 5 and a summary and conclusions in section 6.

\section{Small Exact Response Surface Designs}
Let $N$ represent the number of design points and $K$ represent the number of experimental factors. A design point is an $\mathbf{x}':1\times K$ row-vector. We assume all design factors are scaled to range [-1,1] and so the design space is the $\mathcal{X} = [-1, 1]^K$ hypercube. Let $\mathbf{X}:N\times K$ represent the design matrix. While $\mathcal{X}$ denotes the space of candidate \emph{design points} $\mathbf{x}'$, a \emph{design matrix} $\mathbf{X}$ is a collection of $N$ such design points. Thus, the space of all candidate designs is an $NK$-dimensional hypercube and is denoted:
\begin{equation}
	\label{eq:exactModspace}
	\mathbf{X} \in \bigtimes_{j = 1}^N \mathcal{X} = \bigtimes_{j = 1}^N [-1, 1]^K = [-1, 1]^{NK} = \mathcal{X}^N. 
\end{equation}
We consider the second-order linear model which has $p = {K + 2 \choose 2}$ linear coefficient parameters. In scalar form, the second-order model is written 
$$y = \beta_0 + \sum_{i = 1}^K\beta_ix_i + \sum_{i=1}^{K-1}\sum_{j = i + 1}^K\beta_{ij}x_i x_j + \sum_{i=1}^K \beta_{ii}x_i^2 + \epsilon. $$ 
Let $\mathbf{F}:N\times p$ represent the model matrix with rows given by the expansion vector $\mathbf{f}'(\mathbf{x}'_i) = (1 \quad x_{i1} \quad \hdots \quad x_{i2} \quad x_{i1}x_{i2} \quad \hdots \quad x_{i(K-1)}x_{iK} \quad x_{i1}^2 \quad \hdots \quad x_{iK}^2)$. The model can be written in matrix-vector form as $\mathbf{y} = \mathbf{F}\bm{\beta} + \bm{\epsilon}$ where we impose the standard ordinary least squares assumptions $\bm{\epsilon}\sim \mathcal{N}_N(\mathbf{0}, \sigma^2 \mathbf{I}_N)$ where $\mathcal{N}_N$ denotes the $N$-dimensional multivariate normal distribution. The ordinary least squares estimator of $\bm{\beta}$ is $\hat{\bm{\beta}} = (\mathbf{F}'\mathbf{F})^{-1}\mathbf{F}'\mathbf{y}$ which has variance $\text{Var}(\hat{\bm{\beta}}) = \sigma^2(\mathbf{F}'\mathbf{F})^{-1}$. 
The total information matrix for $\bm{\beta}$, specifically $\mathbf{M}(\mathbf{X}) = \mathbf{F}'\mathbf{F}$, plays an important role in optimal design of experiments---all optimal design objective functions are functions of this matrix.

The practitioner must choose a design from $\mathcal{X}^N$ to implement the experiment in practice. An optimality criterion is used to define which candidate designs $\mathbf{X} \in \mathcal{X}^N$ are `good' designs. An optimization algorithm is required to search $\mathcal{X}^N$ to find the `best', or optimal, design. Thus, an exact optimal design problem is defined by the three components:
\begin{enumerate}
	\item The number of design points $N$ that can be afforded in the experiment.
	\item The structure of the model one wishes to fit (here the second-order model).
	\item A criterion which defines an optimal design. This is a function of $\mathbf{M}(\mathbf{X})$.
\end{enumerate}
In this paper we consider the $D$ and $I$ criteria which are described in the next sub-sections.

\subsection{$D$-optimal Designs}
The $D$-score of an arbitrary design $\mathbf{X}$ is defined as the determinant of the inverse of the information matrix:
\begin{equation}
	D(\mathbf{X}) = N^p|(\mathbf{F}'\mathbf{F})^{-1}| \label{eq:D}
\end{equation}
This yields the definition of the $D$-optimal design $\mathbf{X}^*$,
\begin{equation}
\label{eq:Ddes}
	\mathbf{X}^*  := \argmin_{\mathbf{X} \in \mathcal{X}^N} D(\mathbf{X}).
\end{equation}
Thus, a $D$-optimal design is that design which minimizes the volume of the inverse of the information matrix (which is proportional to $\text{Var}(\hat{\bm{\beta}})$). Because these designs minimize the joint uncertainty in the parameter estimates they are a popular choice for screening experiments, and are also popular because of their ease of computation (e.g. the inverse is not required since it is in the determinant) \citep{rodman}. 

As the scale of $D(\mathbf{X})$ is arbitrary and depends on the number of design points $N$ and number of model parameters $p$, it is customary to scale the $D$-score to the \textit{efficiency} scale
\begin{equation}
	D_{\text{eff}}(\mathbf{X}) = 100 \times \left(\frac{D(\mathbf{X}^*)}{D(\mathbf{X})} \right)^{\frac{1}{p}}. 
\end{equation} 

Similarly, the quality of two designs $\mathbf{X}_1$ and $\mathbf{X}_2$ with respect to the $D$-criterion can be compared relatively as follows
\begin{equation}
	D_{\text{releff}}(\mathbf{X}_1, \mathbf{X}_2) = 100 \times \left(\frac{D(\mathbf{X}_2)}{D(\mathbf{X}_1)} \right)^{\frac{1}{p}}
\end{equation} 
and a relative efficiency over 100 means that design $\mathbf{X}_1$ is of better quality than design $\mathbf{X}_2$.

\subsection{$I$-optimal Designs}
The $I$-criterion is sometimes referred to as an integrated variance criterion. $I$-optimal designs are those that minimize the average prediction variance over $\mathcal{X}$. The scaled prediction variance is often defined as \cite{rodman, jobo1}
\begin{equation}
	\text{SPV}(\mathbf{x}')  = N \mathbf{f}'(\mathbf{x}')(\mathbf{F}'\mathbf{F})^{-1}\mathbf{f}(\mathbf{x}').
\end{equation}
The $IV$-criterion for candidate design $\mathbf{X}$ is defined as
\begin{align}
	IV(\mathbf{X}) &:= \frac{1}{V} \displaystyle\int_{\mathcal{X}} \text{SPV}(\mathbf{x}') \mathbf{dx}' \nonumber \\
	&= \frac{N}{V} \text{tr}\biggl\{(\mathbf{F}'\mathbf{F})^{-1} \displaystyle\int_{\mathcal{X}} \mathbf{f}(\mathbf{x}')\mathbf{f}'(\mathbf{x}')\mathbf{dx}'\biggr\} 
\end{align}
where $V = \displaystyle\int_{\mathcal{X}} \mathbf{dx}'$ yields the volume of the design space $\mathcal{X}$. The $IV$-optimal design $\mathbf{X}^*$ is defined as
\begin{align}
	\mathbf{X}^* &:= \argmin_{\mathbf{X}\in \mathcal{X}^N} IV(\mathbf{X}) \nonumber \\
	&= \argmin_{\mathbf{X}\in \mathcal{X}^N} \frac{N}{V} \text{tr}\biggl\{(\mathbf{F}'\mathbf{F})^{-1} \displaystyle\int_{\mathcal{X}} \mathbf{f}(\mathbf{x}')\mathbf{f}'(\mathbf{x}')\mathbf{dx}'\biggr\}. \label{eq:ivwithn}
\end{align}
The quality of a candidate design $\mathbf{X}$ can be measured via its efficiency
\begin{equation}
	I_{\text{eff}}(\mathbf{X}) = 100 \times \left(\frac{I(\mathbf{X}^*)}{I(\mathbf{X})} \right)
\end{equation} 
and two candidate designs can be compared via relative efficiency in a similar way.

\section{Particle Swarm Optimization}
The problem solved by PSO is $\min_{\mathbf{x} \in \mathcal{X}} f(\mathbf{x}) $
where the feasible region may be a hyperrectangle  $\mathcal{X} = \bigtimes_{k = 1}^K [l_k, u_k] \subset \mathbb{R}^K$. We group the lower bounds and upper bounds into vectors $\mathbf{l}_b = \{l_k\}$ and $\mathbf{u}_b = \{u_k\}$. PSO is a stochastic search and so is not guaranteed to find the global optima of $f$. Note that a particle is simply a candidate solution to the optimization of $f$, or an $\mathbf{x} \in \mathcal{X}$.

\subsection{Basic PSO}
\label{sec:1}
This is the most common version of PSO applied to optimal design problems \cite{chen2, chen1, joseph, lukemire, chen3, chen4, shi1, wong}. Algorithm \ref{alg:bpso} contains a full statement of the Basic PSO algorithm in mathematical notation. Let vectors $\mathbf{x}_i:K\times 1$ and $\mathbf{v}_i:K\times 1$ represent positions and velocities of particle $i = 1, \hdots, S$, respectively, where $S$ is the swarm size (i.e. number of particles). Lines 5 and 6 of the algorithm show that starting positions and velocities are initialized randomly, drawn from a multivariate uniform distribution of dimension $K$ denoted $U_K$. Line 7 shows velocity limiting where parameter $v_k^{\text{max}}$ is a velocity limit typically set to the length of the search space in dimension $k$ (this parameter governs the particle step size). Line 8 shows $\mathbf{p}_{\text{best},i} \in \mathcal{X}$ which represents particle $i$'s personal best location (wrt to fitness on $f$) found in the search space. Line 11 shows $\mathbf{g}_{\text{best}}\in \mathcal{X}$ which represents the swarm's global best position found. The main PSO iteration is a \texttt{while} loop (various stopping criteria may be applied). We've adopted synchronous updates and so there are separate \texttt{for} loops for updating all particle's velocities (lines 14-18) and then moving the particles and updating the fitness knowledge (lines 19-30). Line 22 shows a function \texttt{confine}---the purpose is to check whether particle $i$ left $\mathcal{X}$ at this iteration, and if so it is mapped to the boundary of $\mathcal{X}$ in those dimensions and the corresponding elements of $\mathbf{v}_i$ are set to 0 (this is referred to as the ``Absorbing Wall'' confinement).

\begin{algorithm}[htbp]
\setstretch{1.35}
\caption{The Basic PSO Algorithm - Global Communication Topology}
\label{alg:bpso}
%\texttt{Algorithm} Compete, steps of 8 pt are used for loop indents
  \begin{algorithmic}[1]
    \STATE \textbf{Input:} Objective function $f$, swarm size $S$, search limits $\mathbf{l}_b$ and $\mathbf{u}_b$
    \STATE \textbf{Output:} Particle swarm solution (global best position) $\mathbf{g}_{\text{best}}$
		\STATE \hspace{0pt}{\texttt{//} Initialize the swarms position, velocity, and fitness knowledge}
		\STATE \hspace{0pt}{\texttt{for each }{$i = 1, \hdots, S$ \texttt{\texttt{do}}}} \label{initialize:loop:1}
		\STATE \hspace{16pt}{$\mathbf{x}_i \sim U_K(\mathbf{l}_b, \mathbf{u}_b)$}
		\STATE \hspace{16pt}{$\mathbf{v}_i \sim U_K\left(\frac{\mathbf{l}_b - \mathbf{x}_i}{2}, \frac{\mathbf{u}_b - \mathbf{x}_i}{2} \right)$}
		\STATE \hspace{16pt}{$\{v_{ik}\leftarrow\min\{v_{ik}, v_{k}^\text{max}\}\}$}
		\STATE \hspace{16pt}{$\mathbf{p}_{\text{best},i} \leftarrow \mathbf{x}_i$}
		\STATE \hspace{0pt}\texttt{endfor}
		\STATE \hspace{0pt}{\texttt{//} Set current best fitness}
		\STATE \hspace{0pt}{$\mathbf{g}_{\text{best}} \leftarrow \underset{\mathbf{\mathbf{x}_i \in \{\mathbf{x}_1, \mathbf{x}_2, \hdots, \mathbf{x}_S\}}}{\mathrm{argmin}} f(\mathbf{x}_i)$}
		\STATE \hspace{0pt}{\texttt{//} Main PSO loop}
		\STATE \hspace{0pt}{\texttt{while} stopping criteria not met \texttt{do}} \label{search:loop:2}
		\STATE \hspace{16pt}{\texttt{for each }{$i = 1, \hdots, S$ \texttt{do}}} 
		\STATE \hspace{32pt}{\texttt{//} Update velocities}
		\STATE \hspace{32pt}{$\mathbf{v}_i \leftarrow \omega \mathbf{v_i} + c_1 U_K(\mathbf{0}, \mathbf{1}) \odot\left(\mathbf{p}_{\text{best},i} - \mathbf{x}_i \right) + c_2U_K(\mathbf{0}, \mathbf{1}) \odot\left(\mathbf{g}_{\text{best}} - \mathbf{x}_i \right)$}
		\STATE \hspace{32pt}{$\{v_{ik}\leftarrow\min\{v_{ik}, v_{k}^\text{max}\}\} $}
		\STATE \hspace{16pt}\texttt{endfor}
		\STATE \hspace{16pt}{\texttt{for each }{$i = 1, \hdots, S$ \texttt{do}}} 
		\STATE \hspace{32pt}{\texttt{//} Update positions and fitness}
		\STATE \hspace{32pt}{$\mathbf{x}_i \leftarrow \mathbf{x}_i + \mathbf{v}_i$}
		\STATE \hspace{32pt}{$\mathbf{x}_i \leftarrow \text{confine} (\mathbf{x}_i)$}
		\STATE \hspace{32pt}{\texttt{//} Update fitness and knowledge of $f$ given new positions}
		\STATE \hspace{32pt}{\texttt{if} $f(\mathbf{x}_i) < f(\mathbf{p}_{\text{best},i})$} 
		\STATE \hspace{48pt}{$\mathbf{p}_{\text{best},i} \leftarrow \mathbf{x}_i$}
		\STATE \hspace{48pt}{\texttt{if} $f(\mathbf{p}_{\text{best},i}) < f(\mathbf{g}_{\text{best}})$}
		\STATE \hspace{64pt}{$\mathbf{g}_{\text{best}} \leftarrow \mathbf{p}_{\text{best},i}$}
		\STATE \hspace{48pt}{\texttt{endif}}
		\STATE \hspace{32pt}{\texttt{endif}}
		\STATE \hspace{16pt}\texttt{endfor}
		\STATE \hspace{0pt}{\texttt{endwhile}}
    \RETURN
  \end{algorithmic}
\end{algorithm} 

\begin{figure}%
    \centering
    \subfloat[\centering global communication topology]{{\includegraphics[width=5cm]{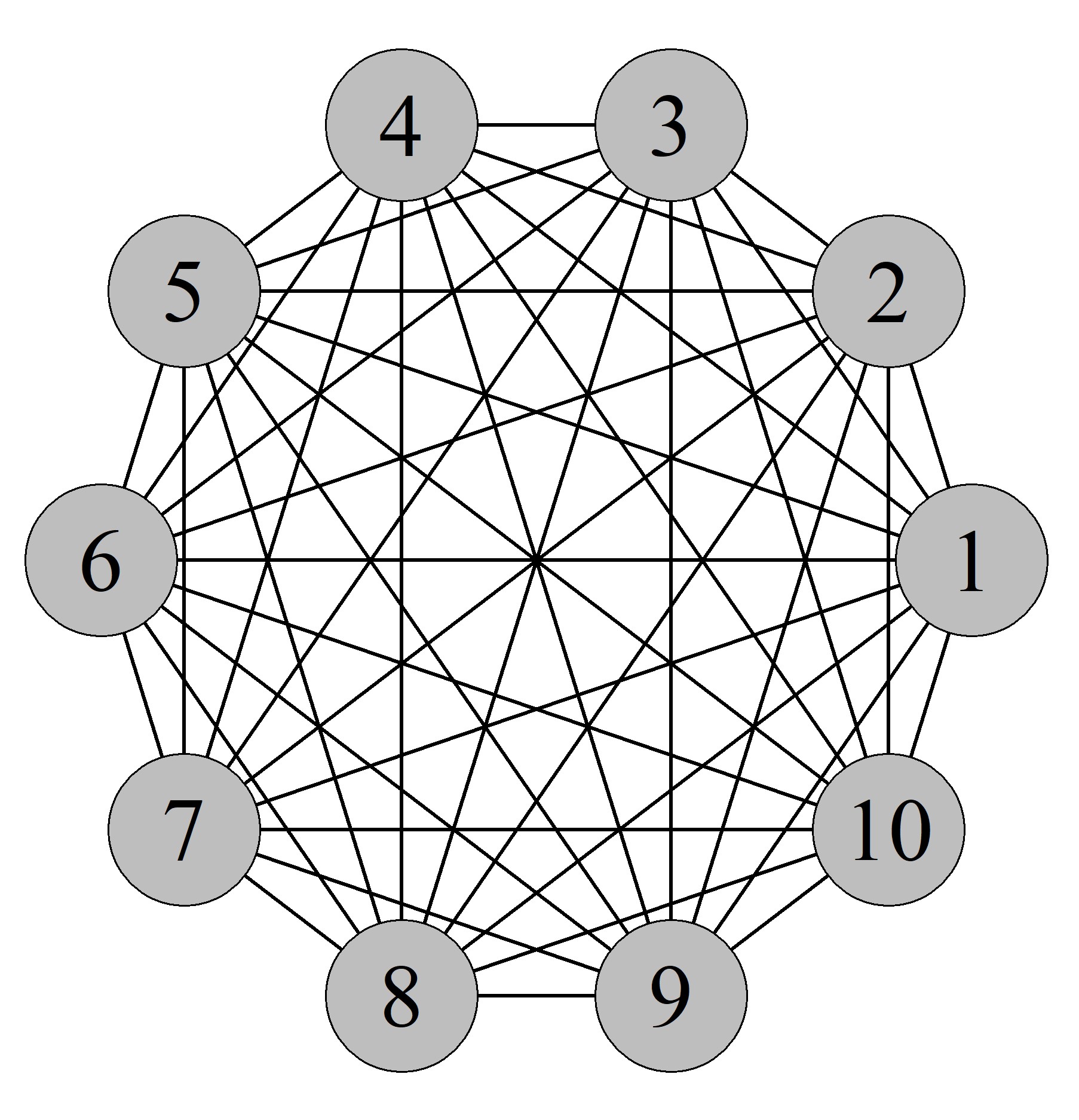} }}%
    \qquad
    \subfloat[\centering local communication topology]{{\includegraphics[width=5cm]{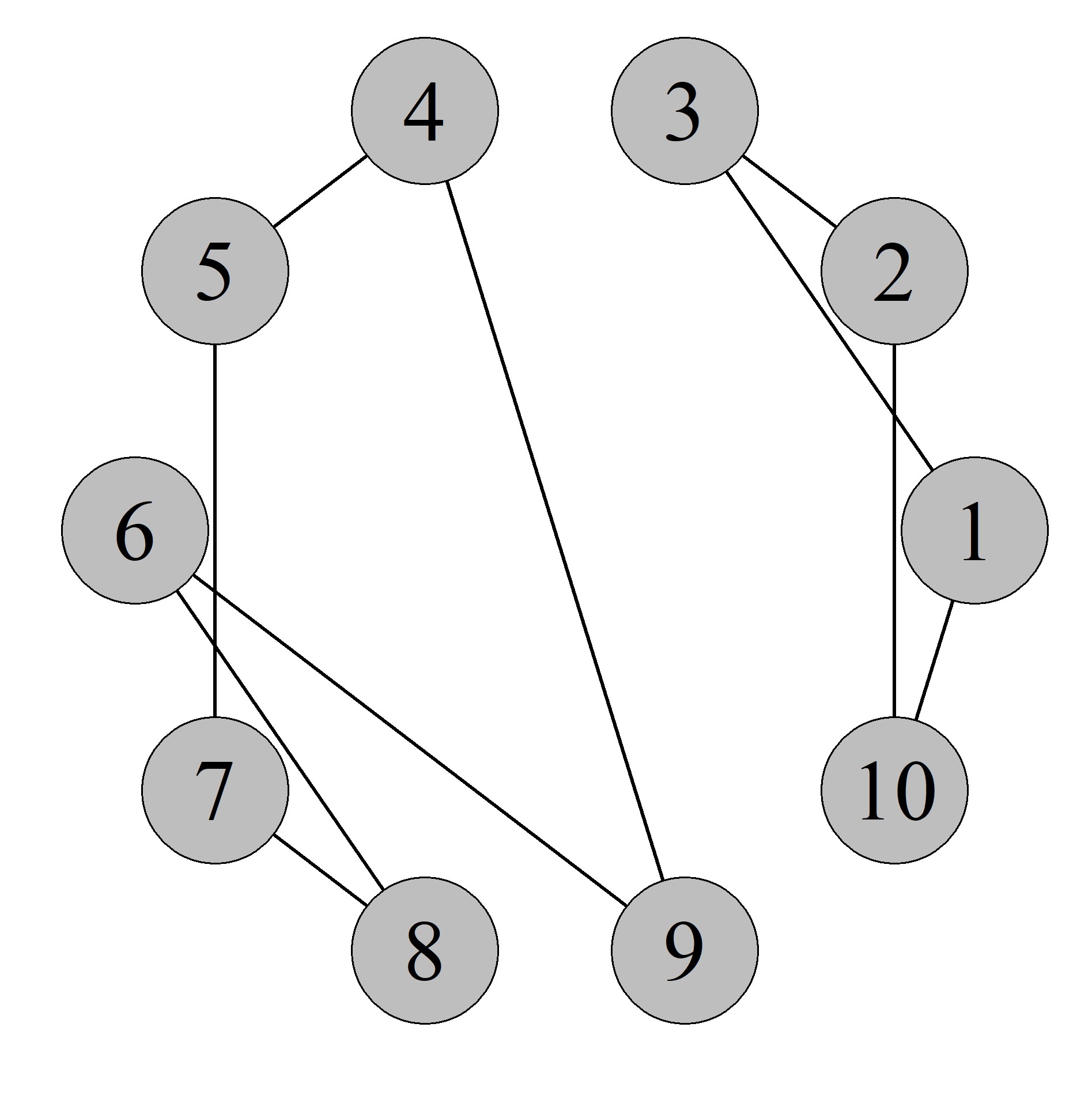} }}%
    \caption{For $S=10$ particles (a) illustration of the global communication topology used in Basic PSO, and (b) illustration of the random local communication topology with 2 links per particle used in SPSO2007.}%
    \label{fig:topos}%
\end{figure}

Swarm communication occurs in the velocity update equation (line 16) which we restate below as a function of iteration index $t$:
\begin{flalign}
  && \mathbf{v}_i(t+1) & = \omega \mathbf{v}_i(t) && \text{(inertia)}   \label{eq:inertia} \\ 
	&&  + & c_1 U_K(\mathbf{0}, \mathbf{1}) \odot\left(\mathbf{p}_{\text{best},i} - \mathbf{x}_i(t) \right) && \text{(cognitive)} \label{eq:cognitive}  \\ 
	&&  + & c_2U_K(\mathbf{0}, \mathbf{1}) \odot\left(\mathbf{g}_{\text{best}} - \mathbf{x}_i(t) \right) && \text{(social)} \label{eq:social} 
\end{flalign}
where $\odot$ represents the Hadamard product (elementwise multiplication). The first term is referred to as inertia because this term is the particle's resistance to changing from its current trajectory. The second term is called the cognitive component as it involves the personal best position $\mathbf{p}_{\text{best},i}$ (the particle's memory) and creates a tendency for the particle to revisit this position. The social component represents swarm communication due to use of the groups best position $\mathbf{g}_{\text{best}}$ and induces a tendency for particles to revisit this point in $\mathcal{X}$. As the value $\mathbf{g}_{\text{best}}$ is known to all particles at each iteration, Basic PSO is said to use the global communication topology. Figure \ref{fig:topos} (a) presents a network diagram illustrating the global communication topology used in Basic PSO.

Optimal values of velocity weightings $\omega, c_1$ and $c_2$ were researched extensively \citep{shi_1998,eberhart_1998,shi2,clerc2,ebershi_3}. For SPSO algorithms these constants are set at 
$$\omega = \frac{1}{\log(2)}, \qquad c_1 = c_2 =\frac{1}{2} +\log(2)$$
and have been demonstrated to balance swarm exploration while promoting convergence, therefore, these quantities do not require further study if the optimization problem resides in the standard Euclidean geometry. These values guarantee that the swarm will eventually settle down to a consensus point for the solution to the optimization.

\subsection{SPSO2007 - Random Local Communication Topology}
This version of PSO has not been applied to optimal design problems and is slightly more costly \cite{spso2_2011}. We will illustrate whether the additional cost is justified for optimal design problems. SPSO2007 initializes the swarm the same way as Basic PSO and also uses the absorbing wall particle confinement. The main difference from Basic PSO is the use of a local communication topology. We present a diagram illustrating an instance of a random local communication topology with $k=2$ communication links in Figure \ref{fig:topos} (b). In SPSO2007 communication links are randomly generated for each particle with \cite{hal_spso} recommends setting the expected number of links at $k=3$. In any iteration the number of communication links for a specific particle can range from 1 (the particle only communicates with itself) to $S$ (the specific particle may communicate with all other particles), but on average each particle will inform 3 other particles. This defines a communication neighborhood for each particle and introduces the term $\mathbf{l}_{\text{best},i} \in \mathcal{X}$ which represents the \textit{local} best position found among the particles in the communication neighborhood of particle $i$. The term $\mathbf{l}_{\text{best},i}$ is substituted for $\mathbf{g}_{\text{best}}$ in the velocity update equation social component in Eq. (3). The fitness at the global best position $f(\mathbf{g}_{\text{best}})$ is still tracked and if it does not improve in an iteration (i.e. the solution stagnates) then the neighborhood links are randomly reinitialized to promote exploration of the swarm. The velocity update is slightly modified if $\mathbf{l}_{\text{best}}
 = \mathbf{p}_{\text{best},i}$ then in this case the social term is dropped from the update for particle $i$. 

On some classes of problems, the random local communication topology has been observed to increase PSOs likelihood of finding the global optimum at the expense of more iterations, \ie the particles explore the space more thoroughly under this topology \cite{engel, topos}. Therefore, we hypothesize that SPSO2007 (local communication topology) may be beneficial over Basic PSO (global communication topology) to generating exact optimal designs as the objective functions in exact optimal design have many local sub-optima.

\subsection{Extending PSO to Generating Optimal Designs}
The formulation of PSO described in previous sections optimizes functions that take vector inputs. Optimal design objectives are functions that take matrix inputs. There exist several long-standing PSO packages and in particular we recommend \texttt{R}'s \cite{hydropso} as it is written by prominent PSO researchers and gives access to the variants of PSO we study in this paper. Transforming the exact optimal design problem to work with existing codes is straigthforward, simply vectorize candidate designs as $\mathbf{x}^*:NK\times1 = \text{Vec}(\mathbf{X})$. This is consistent with the approach of adapting PSO to other optimal designs problems \cite{chen2, chen1, joseph, lukemire, chen3, chen4, shi1, wong}.

\section{Experimental Design and PSO Performance Analysis}

\subsection{Experimental Design}
We implemented a large computer experiment on each of 21 designs scenarios as presented in Borkowski (2003) \cite{jobo1}. Borkowski (2003) used a genetic algorithm (GA) to generate proposed exact optimal designs for the $D$-,  $I$-criterion for the second-order response surface model and for 21 distinct design scenarios over $K = 1, 2$ and $3$ factors. These designs have since been reproduced by several authors, \eg Rodriquez et. al. (2010) and Hernandez and Nachtsheim (2018) \cite{jobo1, rodman, gIopt}. Thus we take the GA generated design as the comparator and published the results of PSO-generated designs as measured by relative efficiency as compared to the GA-generated designs. Borkowski (2003) has become a standard set of data and proposed catalog of exact optimal designs which is highly used by the optimal design community to develop, benchmark, and validate new exact optimal design search algorithms, see \eg \cite{rodman, gIopt, saleh}. The 21 design scenarios we apply PSO to are comprised of $K = 1, 2, 3$ design factors and for experiment sizes $N = 3, 4, 5, 6, 7, 8, 9$, $N = 6, 7, 8, 9, 10, 11, 12$, and $N = 10, 11, 12, 13, 14, 15, 16$ respectively. Our computer experiment is designed to study the effects of three factors on performance of PSO for generating exact optimal designs:
\begin{enumerate}
	\item \textbf{Swarm Size} $S$: levels $S = 50, 150, 500$.
	\item \textbf{PSO Algorithm}, specifically swarm communication topology: levels \{Basic PSO (global communication), SPSO2007 (random local communication)\}
	\item \textbf{Optimal Design Class} sp. objective function: levels $D$- and $I$-criterion. 
\end{enumerate}

We identify the following useful responses:
\begin{enumerate}
	\item (efficiency) Algorithm cost measured as the median number of function evaluations.
	\item (efficacy and efficiency) Probability of a single PSO run to return a highly optimal design, with `highly optimal' considered as a design with 95\% or greater efficiency relative to the best known and published exact design. 
	\item (efficacy) Algorithm success measured as the proportion of runs that return an optimal design as good or better than the best design generated using GA and reported by Borkowski (2003) .
\end{enumerate}

All of our PSO searches were run on an Intel core i7-6700K (with 8 computer cores) running at 4.0Ghz. We implemented Basic PSO and SPSO2007 in the \texttt{Julia} language. We ran $n_{\text{run}} = 140$ independent PSO runs on each design scenario, optimality criterion, PSO algorithm and swarm size yielding a total of 35280 PSO runs implemented in this study. Our stopping criterion was either a non-zero change in the objective score less than the square root of machine epsilon (about $10E$-08), or 100 stagnations of the objective score (i.e. PSO does not find a better solution for 100 iterations).

\subsection{Assessing Efficiency - Analysis of PSO computing cost}
We produced a panel of graphics showing the CPU run-time in seconds for each PSO-version. We analyzed the resulting PSO performance data via linear regression. We take number of function evaluations as the response and model these as a function of $K$, $N$ nested in $K$, $S$, PSO-version (global or local communication topology), and objective class ($D$- or $I$-optimality). A preliminary model suggested that we needed to transform the response to $y = \log (\text{median } \# f \text{ evaluations})$. We found that the first order model fit the data well. Thus, we present results for analysis of the computer experiment data under the following model: 
\begin{align}
 \log(\text{median } \# f \text{ evaluations}) = \beta_0 & + \beta_1K + \beta_2 N(K) + \beta_3 S \nonumber\\
	& + \beta_4 \text{PSO-version} + \beta_5 \text{Objective-class} + \epsilon 
	\label{eq:lmod}
\end{align}
where the errors are assumed to be independent with $\epsilon \sim \mathcal{N}(0, \sigma^2)$. This analysis allows study of the relative impacts of each factor and their interactions on the efficiency of the PSO algorithm.

\subsection{Assessing Efficiency and Efficacy - Analysis of PSO ability to generate highly efficient designs in single runs}
We produced boxplot graphics showing the distribution of the best designs found in the $n_{\text{run}}=140$ PSO runs and plotted the relative efficiency of PSO-generated designs vs. the GA-generated designs of Borkowski (2003) \cite{jobo1}. In addition we computed and reported the proportion of the $n_{\text{run}}=140$ PSO runs (per design scenario) returned a design with $\ge$ 95\% efficiency relative to the best known exact optimal design. If this proportion is high, this is positiive information from the perspective of the practitioner because it means, in contrast to local optimizers such as CEXCH, one would not have to run PSO many thousands of times to find a `good enough' optimal design. Further, if this proportion is hight for small number of particles $S$, then the computation cost for each independent run of PSO is further minimized.

\subsection{Assessing Efficacy - Analysis of PSO ability to reproduce currently known exact optimal deisigns}
\label{sec:apso}
We perform a comprehensive quantitative analysis of the experiment by condensing the data into binary \{success, not success\} by defining `success' as $$\mathcal{S}_r = \{ f_r(\mathbf{G}_{\text{best, PSO}}) \ge f(\mathbf{X}_{\text{GA}})\} $$
where $\mathbf{X}_{\text{GA}}$ is the optimal design found by the GA as reported in \cite{jobo1} for the $r$th run with $r = 1, \hdots, n_{\text{run}} = 140$. Then an estimate of the probability that a single run of the PSO algorithm will find the global optimal design (conditioned on the design scenario, swarm size, and optimality criterion) is
\begin{equation}
\label{eq:psosssss}
	\widehat{\Pr}\{\text{successful PSO run}\} = \frac{\sum_{i=1}^{n_{\text{run}}}I(\mathcal{S}_r)}{n_{\text{run}}}
\end{equation}
where $I(\cdot)$ is the indicator function. Note that this is a very strict criterion, we are estimating the probability that PSO will generate a design at least as efficient as \textit{the currently known optimal design} from 140 PSO runs. With this response we can estimate the effect of each factor, $K$, $N$ nested in $K$, $S$, PSO-version, and objective class, on the probability of PSO finding the global optimal design. To implement this analysis we utilized general additive models (GAMs) and the \texttt{R} packages \texttt{gamlss} and \texttt{gamlss.inf} \cite{R, gamlss, gamlssinf}. The simplest and best fitting model we found for these data is as follows. The pdf for these data with $y$ as the success probability, is represented as
\begin{equation}
 f_Y(y|\bm{\theta}, \xi_1) = \begin{cases} (1-\xi_1)f_B(y|\bm{\theta}) & \text{ if } 0 < y < 1 \\ \xi_1 & \text{ if } y = 1 \end{cases}
\end{equation}
where $f_B$ is a Beta distribution with parameter $\bm{\theta} = (\mu, \sigma)$ representing the mean and variance of the Beta distributions (these are functions of the usual Beta shape parameters), and $\xi_1$ is a point mass which governs the probability inflation of $y=1$ observations. The data are transformed to the logit space $\eta = \text{logit}(y)$ and mean and standard deviation of the Beta distribution are written as functions of the predictor variables via a logit link function: $\text{logit}(\mu)= \eta_1 = \mathbf{X}_1\bm{\beta}_1$ and $\text{logit}(\sigma) = \eta_2 = \mathbf{X}_2\bm{\beta}_2$ where $\mathbf{X}_i$ is the model matrix containing the predictors for each linear function. It is assumed that $\eta \overset{iid}{\sim} \mathcal{N}(\eta_1, \eta_2)$ for the fitting of the linear coefficients. A similar link for the one inflation is required: $\text{logit} (\xi_1) = \eta_3 =  \mathbf{X}_3\bm{\beta}_3$. Last, note that the linear models for $\eta_1$, $\eta_2$ and $\eta_3$ can all contain different functions of the predictors. We fit the one-inflated Beta regression in \texttt{R} packages \texttt{gamlss} and \texttt{gamlss.inf} \cite{R, gamlss, gamlssinf}. After a short model selection exercise we found that, for $\mu$, only the linear terms of each experimental factor were needed. We found that only experimental factors $S$, PSO-version, and Objective were needed in the linear model for $\sigma$ in link space. All factors were found to be important for supporting the one-inflation mass point $\xi_1$. We provide coefficient estimate tables for $\sigma$ and $\xi_1$ in the Appendix as Tables \ref{tab:sighat}1 and \ref{tab:xihat}2, respectively.

Conventional wisdom among the optimal design community is that $K$ (number of factors) is more important than $N$ (number of runs of the practitioner's experiment) in determining the difficulty of the optimal design generation problem---we are able to provide a formal test of this hypothesis via this modeling approach. We also tested for the presence of a PSO-version by objective class interaction, but this interaction was not found to be statistically significant. 

In the next section we provide the results of the analysis of the computer experiment. 

\section{Results}
\begin{figure}%
    \centering
\includegraphics[width=\textwidth]{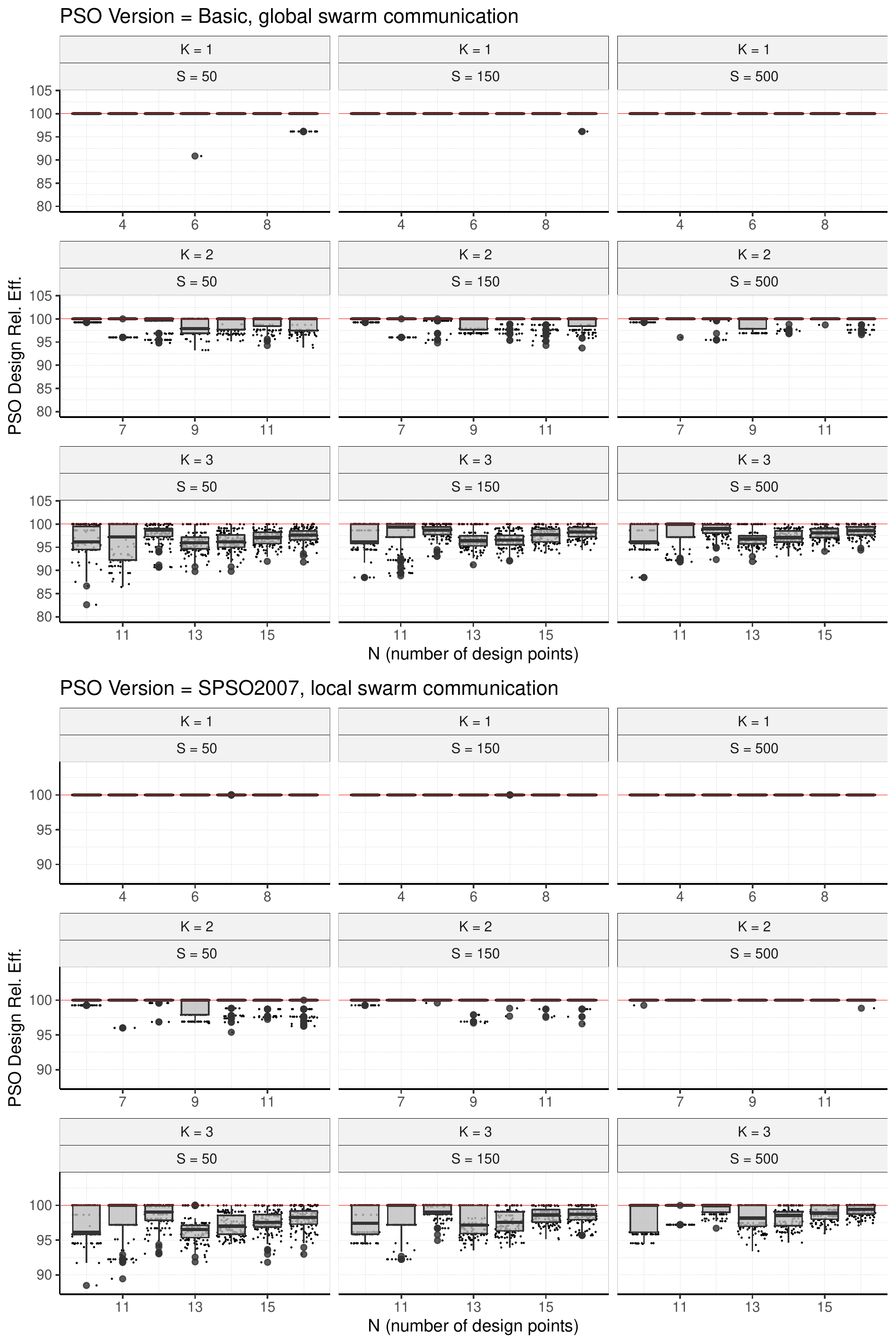} %
    \caption{$D$-optimal designs: PSO generated design efficiencies vs the GA generated designs of Borkowski (2003).}%
    \label{fig:deffs}%
\end{figure}

Figure \ref{fig:deffs} contains the distribution of relative design efficiencies for the PSO searches for $D$-optimal designs as a function of all experimental factors. The data scale on the vertical access is the relative efficiency of the PSO-generated design to the GA-generated design. On this scale, a value of 100 means that PSO generated a design with equal quality to the GA-generated design, and a value over 100 means the PSO design is superior to the GA-generated design. This graphic illustrates that in all cases, PSO was able to find the currently known $D$-optimal design (evidenced by the boxplots hitting 100 on the relative efficiency scale). Further, these graphics show that PSO is guaranteed to produce a design with high efficiency in a single run, evidenced by the boxplots for many of the design scenarios being most dense above, say, 90\% relative efficiency. This is a strong result in support of PSO as a generator for exact optimal designs, \eg Goos and Jones (2011) recommend running the CEXCH algorithm at least 1000 times, and we have run PSO approximately 10\% of that number. Additionally, comparison of the Basic PSO results (global communication topology) to the SPSO-2007 results (local communication topology) indicate that SPSO-2007 and the local communication topology is more efficient at finding better optimal designs, \ie the boxplots are highly dense around 95\% relative efficiency.  These initial results speak well for the ability of PSO to find highly efficient $D$-optimal designs.

\begin{figure}%
    \centering
\includegraphics[width=\textwidth]{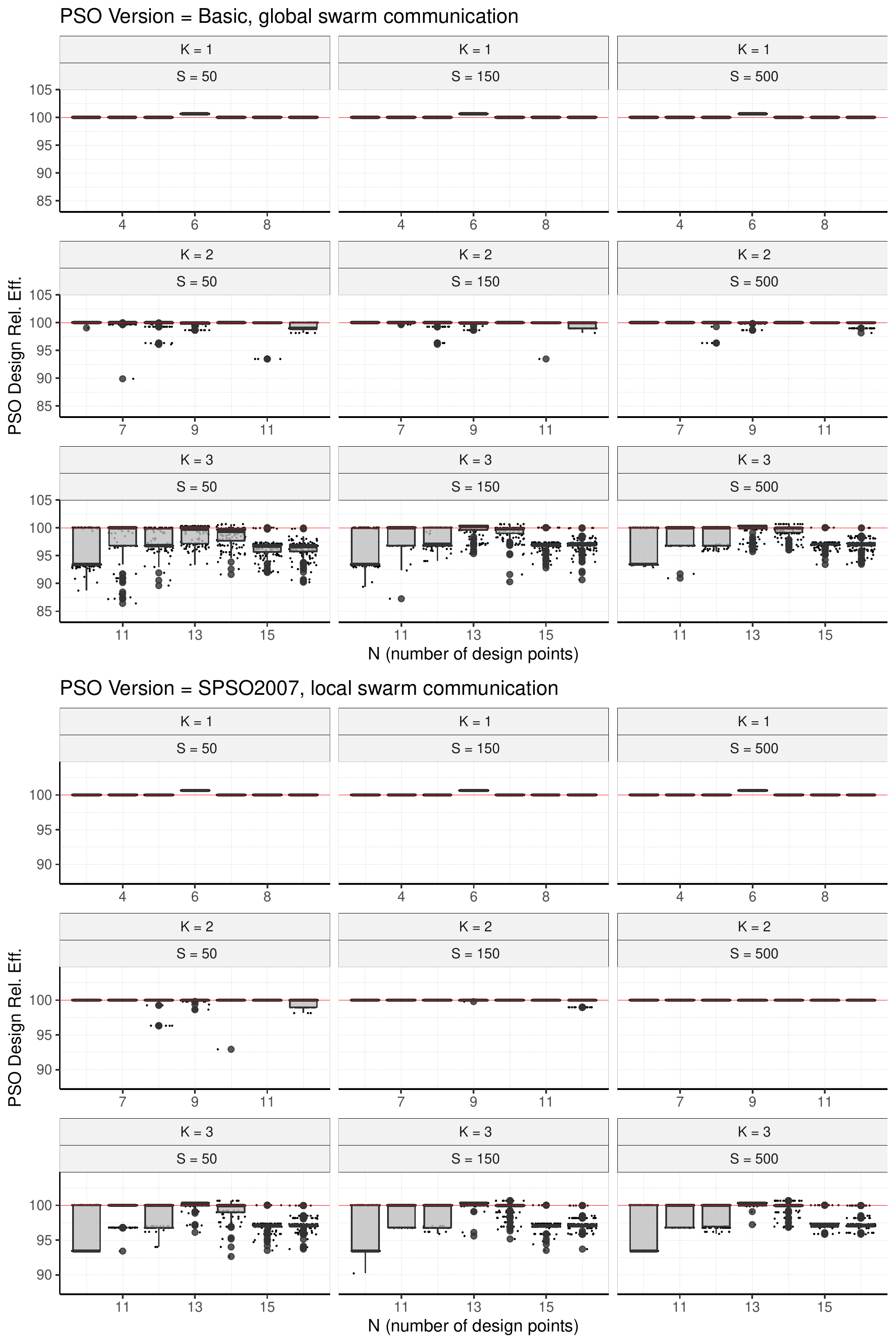} %
    \caption{$I$-optimal designs: PSO generated design efficiencies vs the GA generated designs of Borkowski (2003).}%
    \label{fig:ieffs}%
\end{figure}

Figure \ref{fig:ieffs} contains relative efficiency of the PSO-generated designs vs. the proposed GA design for the $I$-optimal design searches. In this case it can be seen that PSO was able to find a better $I$-optimal design than GA, specifically: $K=1, N=6$, $K=3, N = 13, 14$. In the latter cases the improvement are about 3\% more efficient. These designs are available from the author upon request. The rest of these data indicate similar observations as in the $D$-optimal design scenario: all searches generate designs of high efficiency ($>$90-95\%), and SPSO-2007 seems to generate better solutions than Basic PSO.

\begin{figure}%
    \centering
\includegraphics[width=\textwidth]{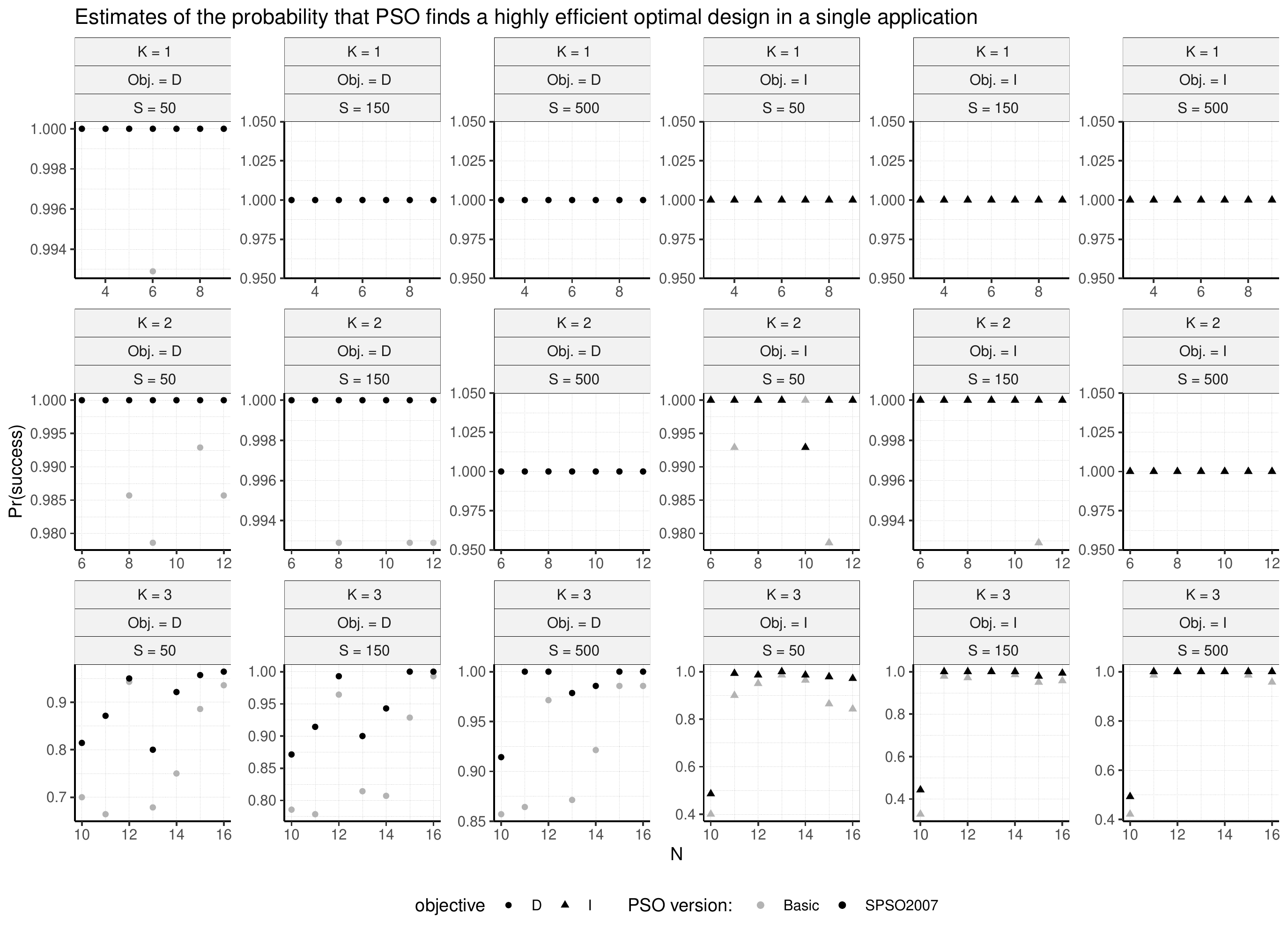} %
    \caption{Estimates of the probability that PSO returns a highly efficient design (one with relative efficiency greater than 95\%) in a single run.}%
    \label{fig:pscss22}%
\end{figure}

Figure \ref{fig:pscss22} reports the proportion of times that PSO was able to produce a design with 95\% relative efficiency or better (comparitor being the best known GA generated design) for each scenario. For either criterion ($D$- or $I$-) and $K = 1,2$ experimental factors, it appears that Basic PSO (global communication topology) generated highly efficient exact designs a high proportion of the time, and SPSO2007 (local communication topology) always generated a design with 95\% or higher relative efficiency, and this is true for any particle size. The implication is that for small response surface scenarios the practitioner can minimized computing cost of a single PSO application by keeping the number of particles in swarm small, and $S = 50$ particles is here evidenced to be highly effective.  The results are slightly different for the $K = 3$ factor design scenarios where, especially for $D$-optimal design searches, one can see the the probability of generating a highly optimal design in a single run drops. Note that the dimension of optimization problems here ranges from $KN = 3*10 = 30$ to $KN = 3*16 = 48$ dimensions, which may be considered optimization searches of moderate size spaces. In this regard SPSO2007 (local communication) offers a clear increase to probability of finding a highly optimal design with many $D$-optimal scenarios Basic PSO having probabilities near 0.7 while SPSO2007 probabilities remain near 0.9. Further, SPSO2007 seems to maintain this efficacy even for the smaller particles swarm sizes $S = 50$ whereas the Basic PSO still does not attain 0.9 probability of generating a highly $D$-optimal design even for large swarms e.g. with $S = 500$ particles. Generally, the $I$-optimal design searches appear easier than $D$-optimal design searches for $K = 3$ factors as evidences by most probabilities of finding a highly efficient design near 1, save for the N = 10 case where the probability of finding a highly $I$-optimal design in a single run drops to 0.5. SPSO2007 and the local communication topology appears to be a better choice than Basic PSO for generating $I$-optimal designs and displays a similar robustness to swarm size, that is, SPSO2007 generates a highly $I$-optimal design with high probability even with small swarm sizes $S = 50$.

Therefore, it is clear that the local communication topology, which has not been explored in optimal design problems to date, is preferable for generating optimal designs an maintains good efficacy and efficiency even for a small number of candidate matrices in the swarm.

\subsection{Effects of experimental factors on algorithm cost}
\begin{figure}%
    \centering
\includegraphics[width=\textwidth]{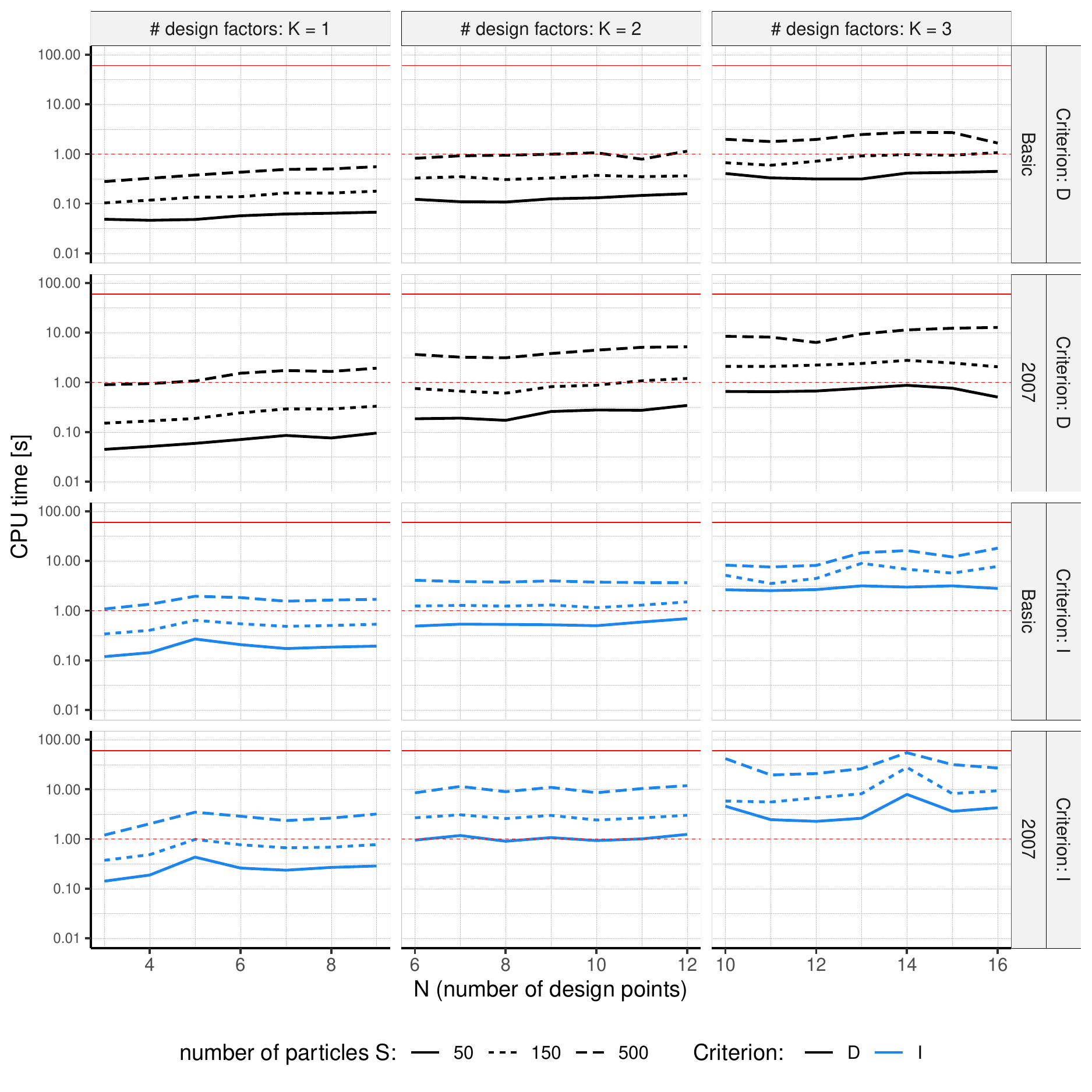} %
    \caption{CPU time in seconds for a single run of the PSO algorithm as a function of the various parameters in this study. The dotted red line marks 1s and the solid red line marks 60s or 1 minute.}%
    \label{fig:cpus}%
\end{figure}

Figure \ref{fig:cpus} contains CPU run time in seconds for a single run of PSO. The graphic illustrates that searching for $I$-optimal designs is slightly more costly than searching for $D$-optimal designs. The biggest cost driver is the number of particles in the swarm $S$, with the $S=500$ particle runs taking roughly ten times longer than $S=50$ particle runs across all other factors (evidenced by the constant distance from the solid lines to the dashed lines). The most expensive search in this suite was for PSO 2007 (local communication topology) for $K=3$ factor designs using $S=500$ particles. However, all cases for this PSO version and number of particles took less than a minute to generate an optimal design. The results are encouraging in the sense that these times are short enough so that PSO is a viable approach to generating optimal designs for experimental practitioners (\eg the GA while superior for generating highly optimal designs is expensive and takes too long to be a useful tool for practitioners \cite{gIopt}). Basic PSO takes about 1s or less to generate a $D$-optimal design for any of the $K=3$ factor design scenarios, no matter the number of particles $S$. This indicates PSO can be run several times in a short time period to search for the global optimal design.

Figure \ref{fig:lofgp} contains graphics showing the log(\# $f$ evaluations) for each design scenario. The graphic clearly shows the increasing impact of $S$ on algorithm cost and indicates that SPSO2007 (local communication topology) appears to be more expensive than Basic PSO (global communication topology), but it is difficult to see the effects of the other experimental factors. Therefore, we now present the analysis of the data under the linear model shown in Eq. \ref{eq:lmod}. The model diagnostics plots for the residuals of the fitted model are provided in Appendix Figure \ref{fig:nfdiags}. These figures indicate that the assumption of constant error variance and normality of the residuals are reasonably met. We provide the coefficient estimates and ANOVA $F$-test in Table \ref{tab:lognfrg}. The overall $F$-test indicates that these predictors are highly significant in explaining variation in the response. The $R^2$ = 0.8903 indicating that 89\% of the variation in log(\# $f$ evaluations) is explained by the regression model. The conditional $t$-tests indicate that each factor is highly significant in their contributions to the model fit with the weakest predictor being $N$ nested in $K$ ($p = 0.0013$). A nice property of a regression model with a log response is that the factor 
\begin{equation}
	\label{eq:respperc}
 100(\exp\{\hat{\beta}_i\}-1)
\end{equation}
is interpreted as ``the percent increase in the response (non-logscale) for a one unit increase in factor $i$"'. Table \ref{tab:estin} contains the quantities shown in Eq. \ref{eq:respperc} with a 95\% confidence interval. The estimates indicate that a one unit increase in the number of design factors $K$ is estimated to increase the expected number of function evaluations by 79.95\% CI: (47.85, 119.03), but may double the number of function evaluations (a 100\% increase is contained in the 95\% confidence interval). Comparatively, a one unit increase in the size of the experiment $N$ for a fixed $K$ is only expected to increase the number of function evaluations by 1.84\%. The conventional wisdom in the optimal design community is that $K$ rather than $N$ contributes more to the difficulty of the problem (even though the dimension of the search is $KN$)---it is nice to see this result quantified. The number of particles is a consistent cost driver with a 100 particle increase in the swarm size expected to increase the cost of the algorithm by 43\%. Next, the coefficient estimate for PSO-version indicates that SPSO-2007 (with local swarm communication topology) is about twice as expensive in number of $f$ evaluations than Basic PSO (global communication topology) CI: (93.7. 140.38). This is consistent with some of the broader PSO literature which describes a tradeoff: the local communication topology is more costly but increases the ability of the swarm to search the space more thoroughly, and therefore is observed to be more robust to local optima entrapment, \eg \cite{topos, engel}. We note that, referring to Figure \ref{fig:cpus}, while SPSO-2007 is twice as costly in \# $f$ evaluations, we are observing wall clock run times for a single PSO search on the order of seconds. Last, the coefficient on the objective-class factor indicates that it is more costly to search for $I$-optimal designs than it is for $D$-optimal designs CI: (56.78, 94.56).

\begin{table}[]
\centering \label{tab:lognfrg}
\caption{Fitted regression model for log($\# f$ evaluations) as a function of the experimental factors.}
\vspace{0.5cm}
\begin{tabular}{lcccc}
\multicolumn{5}{c}{\textbf{Fitted Regression Model}}                                                                                           \\ \cline{1-5}\hline \hline
\textbf{Coefficient} & \textbf{Estimate} & \textbf{Std. Error}  & \textbf{$t$-value}                      & \textbf{Pr(\textgreater{}$|t|$)}   \\ \cline{1-5}
(Intercept)          & 7.5595860         & 0.1102251            & 68.583                                  & \textless 0.0001                   \\
$K$                    & 0.5875167         & 0.0982515            & 5.980                                   & \textless 0.0001                  \\
$K:N$                  & 0.0182658         & 0.0056075            & 3.257                                   & 0.0013                            \\
$S$                    & 0.0043170         & 0.0001399            & 30.859                                  & \textless 0.0001                  \\
psover-SPSO2007       & 0.7690778         & 0.0539799            & 14.247                                  & \textless 0.0001                  \\
objective-I           & 0.5576058         & 0.0539799            & 10.330                                  & \textless 0.0001                  \\
\multicolumn{1}{c}{} &                   &                      &                                         &                                    \\
\multicolumn{5}{c}{\textbf{ANOVA}}                                                                                                            \\ \cline{1-5}\hline \hline
\multicolumn{5}{l}{Residual standard error: 0.4285 on 246 degrees of freedom}                                                                  \\ \cline{1-5}
Multiple R-squared:  & 0.8903            & \multicolumn{1}{l}{} & \multicolumn{1}{l}{Adjusted R-squared:} & 0.888                             \\
\multicolumn{3}{l}{F-statistic: 399.1 on 5 and 246 DF}          & \multicolumn{2}{c}{p-value: \textless 0.0001}                                
\end{tabular}
\end{table}

\begin{table}[]
\centering \label{tab:estin}
\caption{Point estimates and 95\% confidence intervals for the increase in run time measured as \% \# $f$ evaluations for 1-unit increases in the experimental factors.}
\vspace{0.5cm}
\begin{tabular}{ccc}
\multicolumn{3}{c}{\textbf{Expected   Increases in Run-Time as a Percentage of \# $f$ Evaluations}} \\ \hline \hline
\textbf{Coefficient}       & \textbf{Point Estimate}      & \textbf{95\% Confidence Interaval}      \\ \hline
$K$                        & 79.95                        & (47.85, 119.03)                         \\
$K:N$                      & 1.84                         & (0.71, 2.99)                            \\
$S$                        & 0.43                         & (0.41, 0.46)                            \\
psover-SPSO2007            & 115.78                       & (93.7, 140.38)                          \\
objective-I                & 74.65                        & (56.78, 94.56)                         
\end{tabular}
\end{table}

\subsection{Effects of experimental factors algorithm success}
In this analysis success is defined as PSO finding the current best-known optimal design as shown in Eq. \ref{eq:psosssss}.  We now quantify the degree to which each experimental factor impacts success probability via one-inflated Beta regression. 
We can infer the relative impacts of each factor on the probability of generating the global optimal design by interpreting the $\hat{\beta}$'s in Table \ref{tab:mulink}. First, for increasing the number of design points $N$ for a fixed $K$, there is a slight drop in probability of success in logit space $\hat{\beta}_{K:N} = -0.105142$, however, increasing $K$, or the number of design factors, results in a greater drop in logit probability of success $\hat{\beta}_{K} = -0.4988507$, or, increasing $K$ results in a $-0.4988507/-0.105142 = 4.74$  times increase in the difficulty of the optimization problem and a similar increase in $N$. Increasing $S$, the number of particles, increases the probability of success. However, among all factors studied, its impact is quite small: $\hat{\beta}_{S} = 0.0017431$, thus suggesting that one would need to increase the number of particles by more than an order of magnitude (we studied one order of magnitude) in order to receive an appreciable increase in success probability. However, since PSO is here demonstrated to perform well for a small number of particles, this is an encouraging result. The analysis also indicates that the local communication topology has an appreciably large, statistically significant, benefit to probability of finding the global optimum as compared to the global communication topology: $\hat{\beta}_{\text{PSO-version}} = 0.8390914$. Thereby, the local communication topology is highly recommended when implemented PSO to construct exact optimal designs. Last, we learned that generating highly efficient $I$-optimal designs is considerably easier than generating $D$-optimal designs, $\hat{\beta}_{\text{Objective}} = 0.9032175$. We found this to be a somewhat surprising result since the majority of literature for generating exact optimal designs focuses on the $D$-criterion due to its ease of computation. 

\begin{figure}%
    \centering
\includegraphics[width=\textwidth]{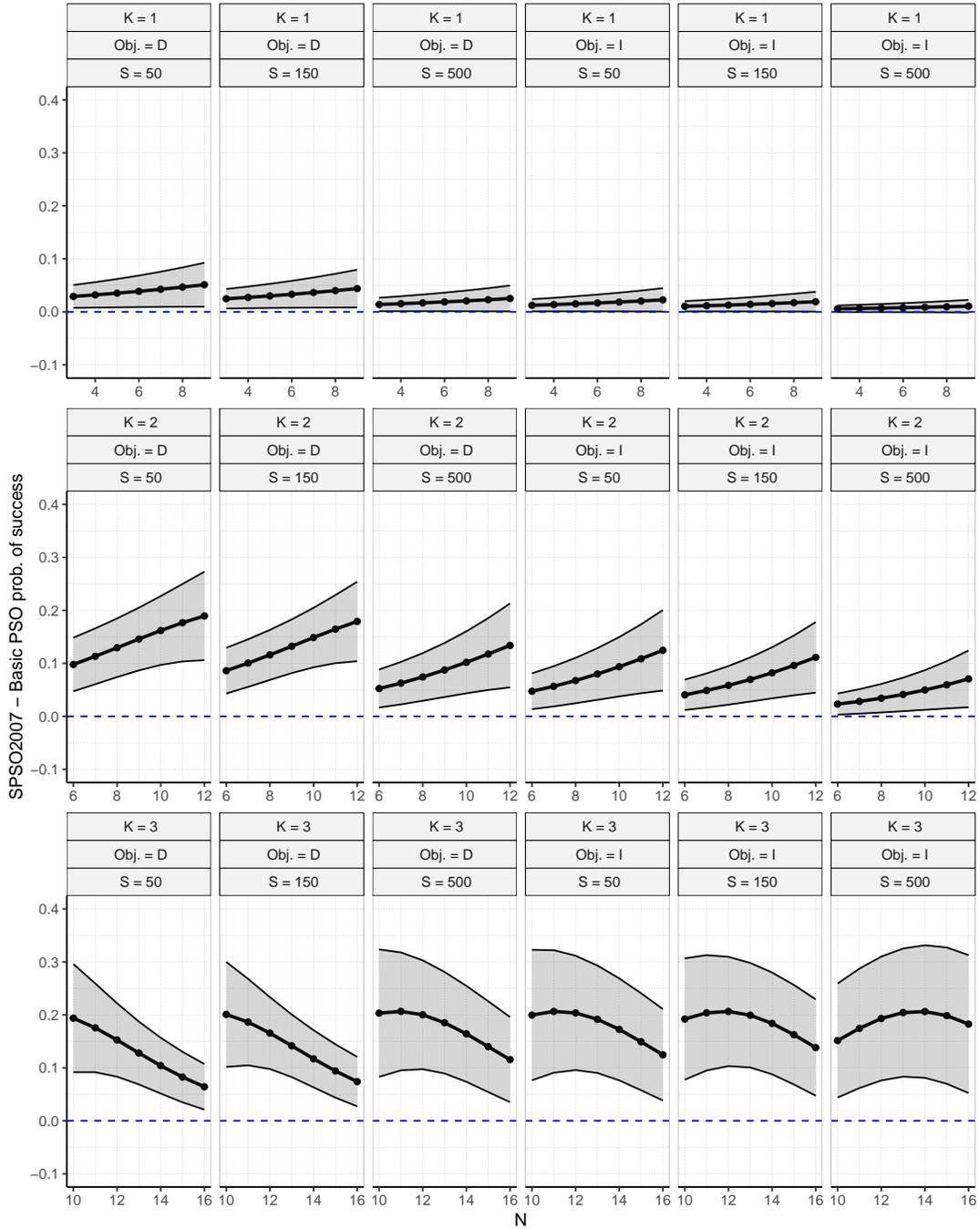} %
    \caption{Estimate of the added benefit of the local swarm communication topology (SPSO2007) in finding the global optimal design as compared to the global communication topology (Basic PSO).}%
    \label{fig:pscss}%
\end{figure}

We use the fitted model to predict the added benefit of using SPSO-2007 with the local communication topology over the Basic PSO version with global communication topology (which was applied in all statistical applications heretofore) by looking at the differences in success probability estimates. We also provide a corresponding 95\% confidence interval on the difference. We present these results in Figure \ref{fig:pscss}. Of course, for the smaller dimension problems (\eg $K=1$) all algorithms and particle sizes could easily find the global optimum, but, there is a measurable improvement by using SPSO-2007 and the local communication topology. As the dimension of the optimization problem increases, we can see that we receive an appreciable benefit by using SPSO-2007 and the local communication topology by realizing a predicted 0.2 increase in probability of success across all $K=3$ designs scenarios. Therefore, these data recommend that the extra cost of using the local communication topology is quite worth the benefit in terms of finding highly-optimal designs more easily with this algorithm. Last, the apparent small effect of the swarm size range we studied can be observed in this graphic. We found this to be a surprising result. However, it also means that a moderate number of particles, $S\approx 100$ is effective for generating highly optimal designs. Note that it appears that the benefit of SPSO-2007 and the local communication topology tends to decrease with the swarm size. This is likely due to the fact that the Basic PSO performs better with more samples of the design space.

\begin{table}[]
\label{tab:mulink}
\caption{Coefficient estimates for the linear function via link on the mean of the Beta distribution.}
\vspace{0.5cm}
\centering
\begin{tabular}{ccccc}
$\mu$                 & \multicolumn{2}{c}{\textbf{link function:}}  & logit                &                     \\
\multicolumn{5}{c}{\textbf{Fitted Linear   Structure}}                                                            \\ \hline \hline
\textbf{Coefficient}  & \textbf{Estimate}     & \textbf{Std.Error}   & \textbf{$t$-value}   & \textbf{Pr($>|t|$)} \\ \hline
(Intercept)           & 3.6228395             & 0.3647829            & 9.931                & \textless{}0.0001   \\
K                     & -0.4988507            & 0.2537637            & -1.966               & 0.0505              \\
K:N                   & -0.105142             & 0.0126353            & -8.321               & \textless 0.0001    \\
S                     & 0.0017431             & 0.0003729            & 4.674                & \textless 0.0001    \\
psover-SPSO2007       & 0.8390914             & 0.1367212            & 6.137                & \textless 0.0001    \\
objective-I           & 0.9032175             & 0.1611937            & 5.603                & \textless 0.0001    \\
                      &                       &                      &                      &                     \\
\multicolumn{5}{c}{\textbf{Deviance Table}}                                                                       \\ \hline \hline
\multicolumn{3}{c}{Number of   observations in fit:}                 & \multicolumn{1}{l}{} & 252                 \\ \hline
\multicolumn{3}{c}{Degrees of freedom   for the fit:}                & \multicolumn{1}{l}{} & 16                  \\
\multicolumn{3}{c}{Residual degrees of   freedom:}                   & \multicolumn{1}{l}{} & 236                 \\
                      &                       &                      &                      &                     \\
\multicolumn{2}{c}{\textbf{Global deviance:}} & -147.6052            &                      &                     \\
                      & AIC:                  & -115.6052            &                      &                     \\
                      & SBC:                  & -59.13429            &                      &                    
\end{tabular}
\end{table}

\section{Conclusions}
In this paper we presented a needed application of PSO to the exact optimal design problem. Further, we conducted the first study of the SPSO2007 and local communication topology for its efficiency and efficacy in generating optimal designs. PSO has been demonstrated to find all currently known exact $D$-optimal designs in a moderate number of runs ($n_{\text{run}}=140)$, and with high probability. This is in contrast to the popular CEXCH which is a well-known local optimizer and, because it is based on a randomly drawn starting candidate, has essentially probability 0 of finding the globally optimal design. PSO found all best known exact $I$-optimal designs, and, for several design scenarios PSO has produced better exact $I$-optimal designs. The results we produced in this paper suggest that, for generating exact optimal designs:
\begin{enumerate}
	\item SPSO2007 with the local communication topology costs twice as much as Basic PSO and the global communication topology in respect to number of function evaluations, however, the wall-clock CPU time for a single run of SPSO2007 is still on the order of seconds.
	\item SPSO2007 is demonstrated to give consistent success in finding highly optimal exact designs even for small swarm sizes, with $S = 50$ being more than adequate for these design scenarios.
	\item SPSO 2007 is estimated to give a 0.2 increase in probability of finding the global optimal design for higher dimensional searches in a single run of PSO.
\end{enumerate}
Therefore the use of  SPSO2007 with the local communication topology is highly recommended over Basic PSO as a tool for generating optimal designs. Further, we expect this benefit to be highly realized as PSO attains more application to higher $K$ experiment sizes.

These results indicate that $K$, \ie the number of design factors, impact the difficulty of the optimal design problem much more than does increasing the number of design points $N$. Specifically, increasing $K$ increases the difficulty of the optimization problem by about 5 times compared to a one unit increase in $N$. This is not a result that has been formally tested and quantified in prior studies.

We also found that generating $I$-optimal designs is less difficult but more expensive than generating $D$-optimal designs. This is evidenced by the higher probability of success curves for $I$-optimal designs shown in Figure \ref{fig:pscss} for the $I$-optimal designs take about 75\% more run time than do searches for $D$-optimal designs.  

In summary, these results support very well the efficiency and efficacy of PSO as a tool for generating exact optimal designs both for the statistics researcher and the applied practitioner of experiments. The cost is here demonstrated to be small enough that PSO can be applied in real time to generate candidate optimal designs, and these data suggest that PSO has a high probability of generating a design with high efficiency in a single run. Ultimately, based on the results of this study, we highly recommend that experimental practitioners add Particle Swarm Optimization, and specifically SPSO2007, to their tool box for generating candidate exact optimal designs.

\bibliographystyle{tfnlm}
\bibliography{mybib}

\section{Appendices}

%\processdelayedfloats %%% See above for an explanation of why this command might be needed here.

\appendix

\section{Regression diagnostics for model on log($\# f$ evaluations)}
\begin{figure}%
    \centering
\includegraphics[width=\textwidth]{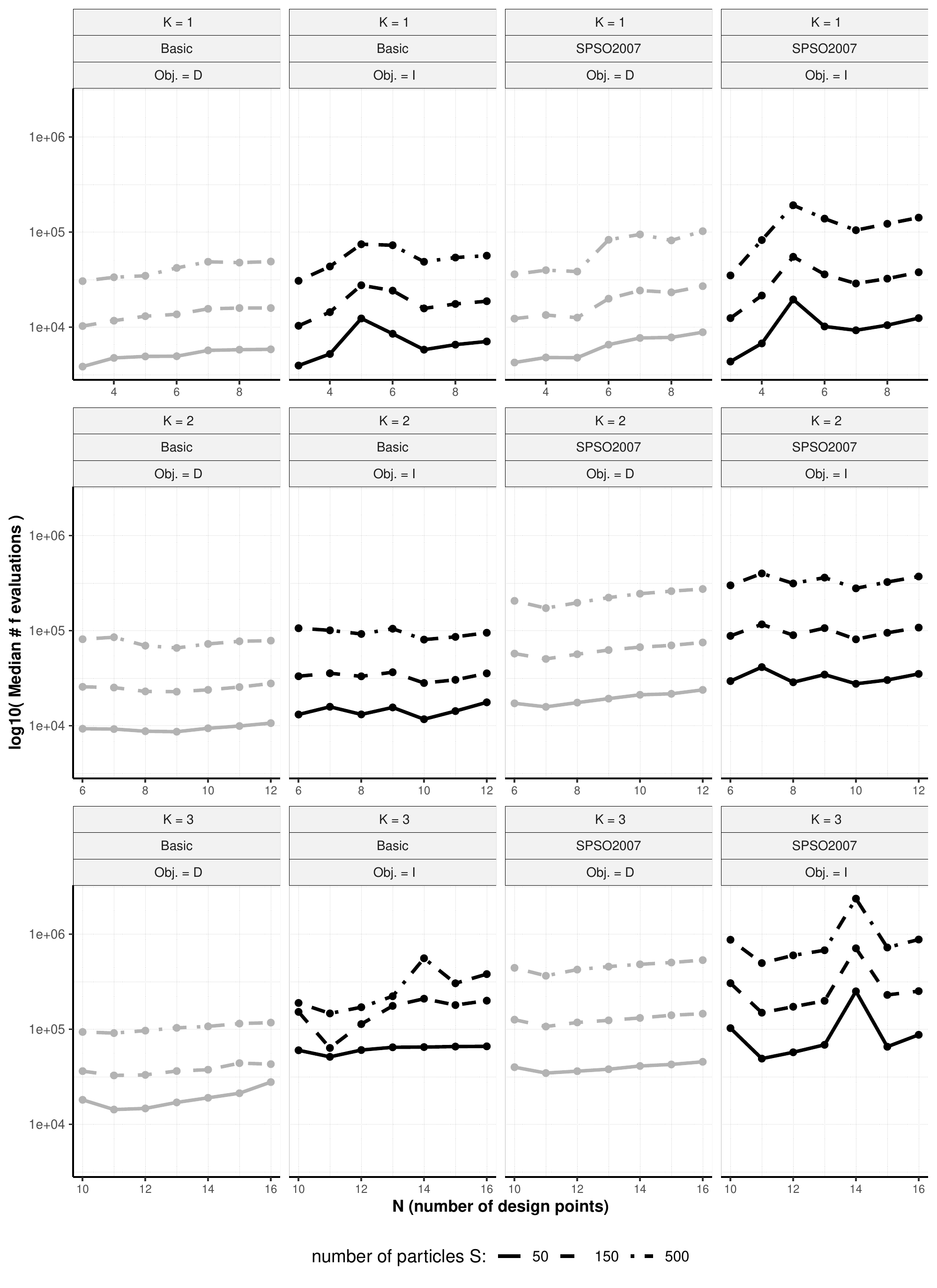} %
    \caption{Median number of $f$ evaluations (log-scale) of the PSO search for the indicated design scenario.}%
    \label{fig:lofgp}%
\end{figure}

Figure \ref{fig:lofgp} shows the data used in the linear model analysis of Eq. \ref{eq:lmod}.

\begin{figure}%
    \centering
\includegraphics[width=\textwidth]{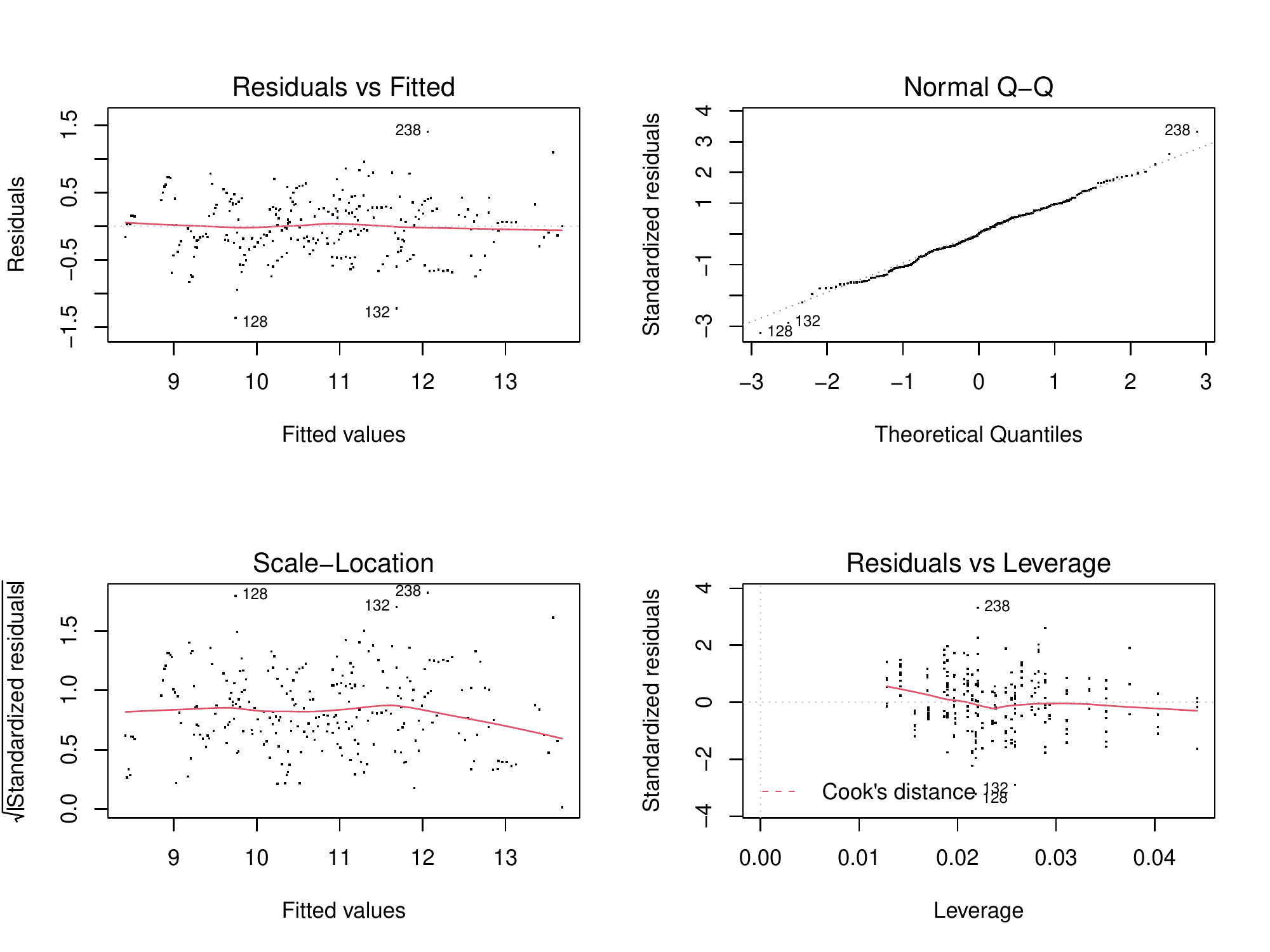} %
    \caption{Standard model diagnostics plots for the regression of log($\# f$ evaluations) on the experimental factors $K$, $N$ in $K$, $S$, PSO-version, and objective. In total these plots indicate that the assumptions of the linear model is met. Therefore, the inferences and uncertainty quantification drawn from this model are on sound footing.}%
    \label{fig:nfdiags}%
\end{figure}

The model diagnostics plots for the model in Eq. \ref{eq:lmod} are provided in Figure \ref{fig:nfdiags}. The residuals vs. fitted and scale-location plot indicate that the constant errors assumption is reasonable for these data. The normal-QQ plot indicates that the normality of residuals assumption is reasonable.

\section{One-inflated Beta regression output and diagnostics on probability that PSO generates the globally optimal design}
\begin{figure}%
    \centering
\includegraphics[width=\textwidth]{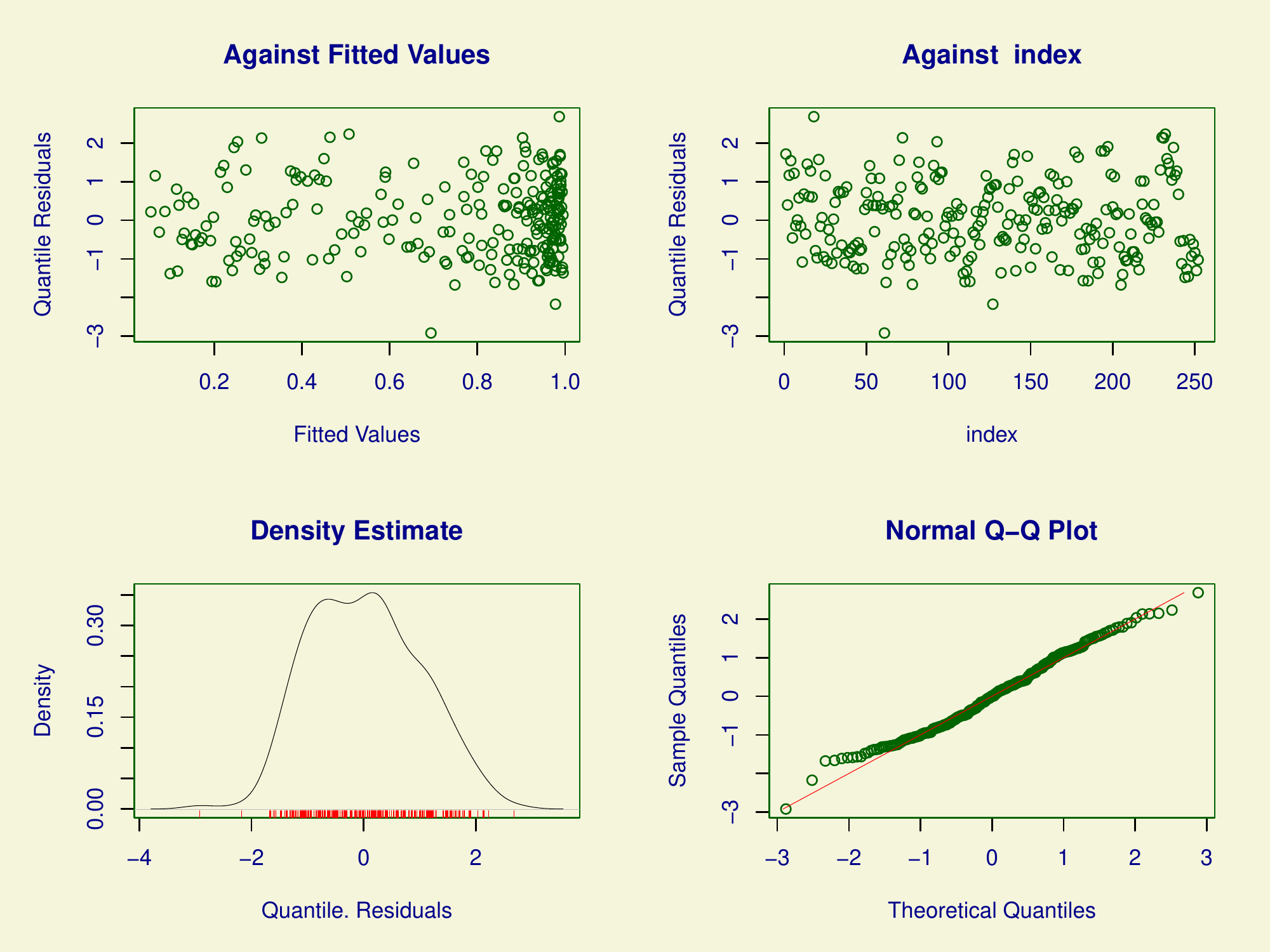} %
    \caption{Model diagnostics in the link function space for the one-inflated Beta regression of the predictors on the probability that PSO generates the globally optimal design.}%
    \label{fig:psdiags}%
\end{figure}

Figure \ref{fig:psdiags} contains the model residuals in link space for the analysis under the model discussed in Section \ref{sec:apso}. The residual plots indicate the normality of residuals and constant error variance are reasonable. Tables \ref{tab:sighat}1 and \ref{tab:xihat}2 contain the GAMLSS estimates of the linear functions in logit space for parameters $\eta_2$ and $\eta_3$, respectively.

\begin{table}[]
\label{tab:sighat}
\caption{Coefficient estimates for the linear function via link on the variance of the Beta distribution.}
\vspace{0.5cm}
\centering
\begin{tabular}{ccccc}
$\sigma$             & \multicolumn{2}{c}{\textbf{link function:}} & logit              &                     \\
\multicolumn{5}{c}{\textbf{Fitted Linear   Structure}}                                                        \\ \hline \hline
\textbf{Coefficient} & \textbf{Estimate}    & \textbf{Std.Error}   & \textbf{$t$-value} & \textbf{Pr($>|t|$)} \\ \hline
(Intercept)          & -1.2169121           & 0.1455171            & -8.363             & \textless 0.0001    \\
S                    & 0.0007124            & 0.0003396            & 2.097              & 0.037               \\
psover-SPSO2007      & 0.2834038            & 0.1473129            & 1.924              & 0.0556              \\
objective-I          & 0.9021098            & 0.1543739            & 5.844              & \textless 0.0001   
\end{tabular}
\end{table}

\begin{table}[]
\label{tab:xihat}
\caption{Coefficient estimates for the linear function via link on the point mass for observing 1's in the one-inflated Beta distribution.}
\vspace{0.5cm}
\centering
\begin{tabular}{ccccc}
$\xi_1$              & \multicolumn{2}{c}{\textbf{link function:}} & logit              &                     \\
\multicolumn{5}{c}{\textbf{Fitted Linear   Structure}}                                                        \\\hline \hline
\textbf{Coefficient} & \textbf{Estimate}    & \textbf{Std.Error}   & \textbf{$t$-value} & \textbf{Pr($>|t|$)} \\ \hline
(Intercept)          & 6.991654             & 1.55294              & 4.502              & \textless 0.0001    \\
K                    & -4.616875            & 1.362098             & -3.39              & 0.0008              \\
K:N                  & -0.256885            & 0.094432             & -2.72              & 0.007               \\
S                    & 0.007486             & 0.002081             & 3.598              & 0.0004              \\
psover-SPSO2007      & 3.058031             & 0.815779             & 3.749              & 0.0002              \\
objective-I          & 4.168887             & 0.953944             & 4.37               & \textless 0.0001   
\end{tabular}
\end{table}

\end{document}